\documentclass[10pt,twocolumn,letterpaper]{article}
\usepackage{cvpr}
\usepackage{times}
\usepackage{epsfig}
\usepackage{xcolor}
\usepackage{graphicx}
\usepackage{color}
\usepackage{enumerate}
\usepackage{amsmath}
\usepackage{amssymb}
\usepackage{multirow}
\graphicspath{{./fig/}}
\usepackage{subcaption}
\usepackage{pifont}
\newcommand{\xmark}{\text{\ding{55}}}
\newcommand{\cmark}{\text{\ding{51}}}

\usepackage[pagebackref=true,breaklinks=true,letterpaper=true,colorlinks=true,bookmarks=false,citecolor=green]{hyperref}

\cvprfinalcopy 

\def\cvprPaperID{4557} 

\ifcvprfinal\fi
\hypersetup{draft} 
\begin{document}

\title{Dually Supervised Feature Pyramid for Object Detection and Segmentation}


\author{\;\; Fan Yang$^{1}$  \;\;   Cheng Lu$^{2}$\;\; Yandong Guo$^{2}$ \;\; Longin Jan Latecki$^{1}$ \;\; Haibin Ling$^{3}$ \\
	{\normalsize $^{1}$Department of Computer and Information Sciences, Temple University, Philadelphia, USA}\\
	{\normalsize $^{2}$Xmotors.ai, Mountain View, CA,  USA} \\
	{\normalsize $^{3}$Department Computer Science, Stony Brook University, Stony Brook, NY, USA.}\\
    {\tt\small \{fyang,latecki\}@temple.edu, \{luc,guoyd\}@xiaopeng.com, hling@cs.stonybrook.edu}
}
\maketitle

\begin{abstract}
   Feature pyramid architecture has been broadly adopted in object detection and segmentation to deal with multi-scale problem. However, in this paper we show that the capacity of the architecture has not been fully explored due to the inadequate utilization of the supervision information. Such insufficient utilization is caused by the supervision signal degradation in back propagation. Thus inspired, we propose a dually supervised method, named \emph{dually supervised FPN} (DSFPN), to  enhance the supervision signal when training the feature pyramid network (FPN). In particular, DSFPN is constructed by attaching extra prediction (\ie, detection or segmentation) heads to the bottom-up subnet of FPN. Hence, the features can be optimized by the additional heads before being forwarded to subsequent networks. Further, the auxiliary heads can serve as a regularization term to facilitate the model training. In addition, to strengthen the capability of the detection heads in DSFPN for handling two inhomogeneous tasks, \ie, classification and regression, the originally shared hidden feature space  is separated by decoupling classification and regression subnets. 
   To demonstrate the generalizability, effectiveness, and efficiency of the proposed method, DSFPN is integrated into four representative detectors (Faster RCNN, Mask RCNN, Cascade RCNN, and Cascade Mask RCNN) and assessed on the MS COCO dataset. 
   Promising precision improvement, state-of-the-art performance, and negligible additional computational cost are demonstrated through extensive experiments. Code will be provided.
   
       
\end{abstract}

\section{Introduction}
\begin{figure}[t]
	\begin{center}
	\includegraphics[width=\linewidth]{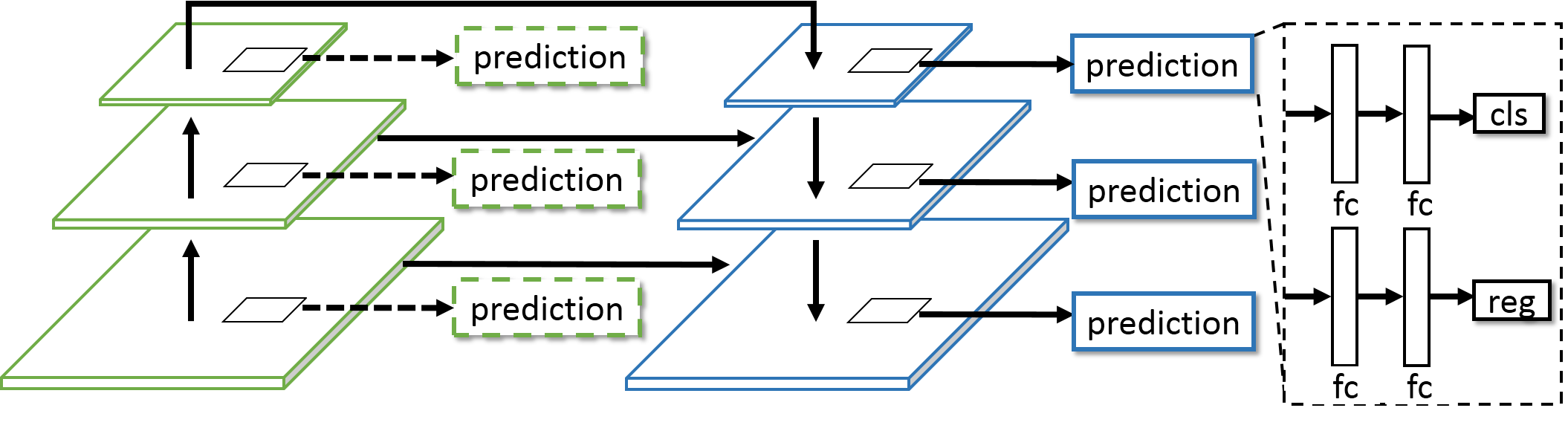}
	\end{center}
	\vspace{-0.3cm}
	\caption{ The diagram of the proposed dually supervised FPN~\cite{lin2017feature} (DSFPN). The green and blue pyramid represent bottom-up and top-down subnets, respectively. 
The green dashed boxes are auxiliary prediction (\ie, detection or segmentation) heads. The black dashed box is the decoupled \textit{cls-reg} heads in detection task. Note that the prediction heads with the same color share weights.}
	\label{fig:dsfpn_framework}
\end{figure}

\begin{figure*}[t]
	\centering
	\begin{subfigure}[b]{0.095\linewidth}
		\includegraphics[width=\linewidth]{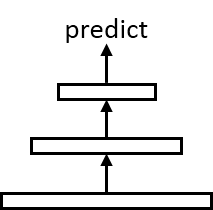}
	\caption{}
		\label{fig:single_head}
	\end{subfigure}
	~
	\begin{subfigure}[b]{0.12\linewidth}
		\includegraphics[width=\linewidth]{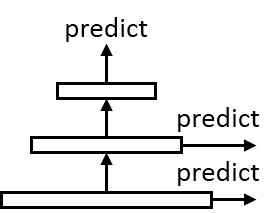}
			\caption{}

		\label{fig:multi_head}
	\end{subfigure}
	~
	\begin{subfigure}[b]{0.23\linewidth}
		\includegraphics[width=\linewidth]{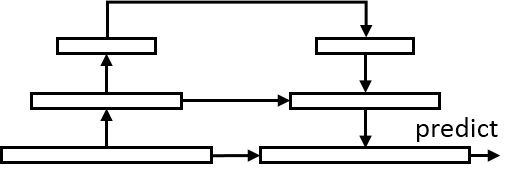}
			\caption{}
		\label{fig:u_single_head}
	\end{subfigure}
	~
	\begin{subfigure}[b]{0.23\linewidth}
		\includegraphics[width=\linewidth]{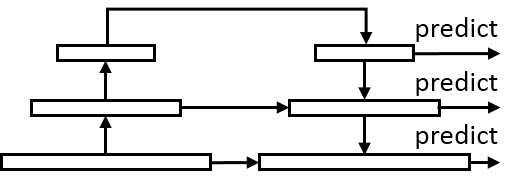}
			\caption{}

		\label{fig:u_multi_head}
	\end{subfigure}
	~
	\begin{subfigure}[b]{0.23\linewidth}
		\includegraphics[width=\linewidth]{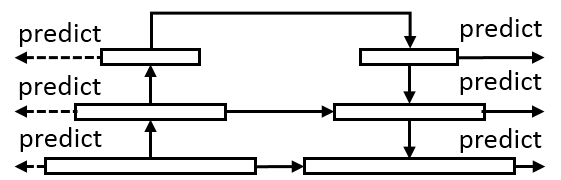}
			\caption{}

		\label{fig:dsfpn}
	\end{subfigure}
	\caption{(a-c) Typical architectures in object classification, detection, and segmentation. (d) The proposed DSFPN. }
	\label{fig:typical_architecture}
\end{figure*}

As one of the most fundamental problems in computer vision, object detection has been studied for decades. Following the rise of deep learning in computer vision, object detection has been greatly advanced in recent years. Numerous deep detectors are proposed and achieve promising results on the large-scale benchmark MS COCO~\cite{lin2014microsoft}. In general, these deep detectors can be categorized into two types: one-stage model~\cite{liu2016ssd,redmon2016you,redmon2017yolo9000,zhang2018single,law2018cornernet,duan2019centernet,yang2019reppoints} and two-stage model~\cite{girshick2014rich,girshick2015fast,ren2015faster,dai2016r,he2017mask,cai2018cascade}. Compared with one-stage approaches, two-stage detection models usually consist of three components: {\it backbone network}, {\it region proposal head} and {\it detection head}. Given object proposals, a feature pyramid architecture (\eg, FPN~\cite{lin2017feature}) is often adopted as the backbone network in the two-stage detectors. After that, a detection head is  attached to the backbone in a lateral connect manner for achieving object localization and classification, as shown in the blue box in Fig.~\ref{fig:dsfpn_framework}. As a natural extension, this framework has also been widely utilized for instance segmentation by adding segmentation heads.

Despite excellent performance obtained by this pyramid architecture, its potential has not been fully explored, owing to the insufficient utilization of supervision information. The reasons of the insufficient utilization can be summarized into two aspects: gradient degradation in feature pyramid and task inconsistency in detection head. In particular, 
the feature pyramid architecture is composited by two components: bottom-up subnet and top-down subnet; the prediction heads are mounted on top-down path (see Fig.~\ref{fig:dsfpn_framework}). This may lead to the inadequate training of bottom-up subnet due to the presence of gradients degradation~\cite{he2016deep}.
Consequently, the top-down feature pyramid, which is composed of features from bottom-up counterpart, has shortage of representation capability. Further, the detection and segmentation performance are not saturated under feature pyramid architecture.

The task inconsistency is an inherent problem~\cite{cheng2018revisiting} in current structure of the detection head, which is constructed by two weights shared sibling subnets for classification and regression, respectively. This structure limits the ability of the head to utilize supervision information. 
Specifically, the sibling subnets deal with the two {\it inconsistent} tasks (classification and regression) with shared FC layers, resulting in sub-optimal solutions for each one~\cite{cheng2018revisiting}. Thus, in feature space the tangled supervision information in the two tasks cannot be effectively leveraged to facilitate the feature learning in the feature pyramid. As a consequence, the detection performance is seriously under-explored.


 Addressing the aforementioned issues, we first propose a novel supervised learning strategy for feature pyramid network, called \textit{dually supervised feature pyramid network} (DSFPN), which is embodied by mounting extra prediction heads (\ie, detection or segmentation) on the bottom-up layers through lateral connection. This design strategy encourages the supervision signal to directly propagate to the bottom-up feature pyramid, so as to strengthen supervision information for the feature learning in it. Moreover, the prediction heads on the bottom-up part can serve as a regularization term to promote the model training. In the test phase, the auxiliary prediction heads are discarded, thus no extra computational cost is needed for model inference.
In DSFPN, the \textit{cls} and \textit{reg} subnets (black dashed box of Fig.~\ref{fig:dsfpn_framework}) are decoupled to enhance the capacity of detection head to handle the task inconsistency, \ie, classification vs regression. Specifically, the decoupling increases the capacity of the heads such that the supervision information can be fully exploited to certain extent to strengthen the guidance of feature learning in the pyramid architecture.

The proposed \textit{dual supervision} (DS) and \textit{decouple} (DC) operations are general to a series of feature pyramid based detectors. The two operations are evaluated by incorporating into four representative FPN-based detectors (Faster RCNN~\cite{ren2015faster}, Mask RCNN~\cite{he2017mask}, Cascade RCNN~\cite{cai2018cascade}, and Cascade Mask RCNN~\cite{cai2018cascade}) with various backbone networks (ResNet-50/101~\cite{he2016deep}, ResNeXt-101~\cite{xie2017aggregated}) on the MS COCO dataset~\cite{lin2014microsoft}. 

In summary, the paper makes the following main contributions:
\begin{enumerate}
	\item A novel DSFPN is proposed to solve the gradient degradation problem in FPN~\cite{lin2017feature} for object detection and segmentation, so as to facilitate feature learning, promote model training. The task inconsistency in detection head is alleviated by decoupling the \textit{cls} and \textit{reg} subnets.
	
	
	
	\item The proposed method consistently improves Faster RCNN~\cite{ren2015faster}, Mask RCNN~\cite{he2017mask}, Cascade RCNN~\cite{cai2018cascade}, and Cascade Mask RCNN~\cite{cai2019cascade} by a promising margin and achieves the state-of-the-art performance on COCO~\cite{lin2014microsoft} with negligible extra computational cost.	
\end{enumerate}

The rest of the paper is organized as follows. Section~\ref{relatedwork} briefly reviews the relevant works. The proposed method is described in detail in Section~\ref{method}. Section~\ref{experiment} presents the experimental results. Section~\ref{conclusion} draws conclusions.


\begin{figure*}[t]
	\centering
	\begin{subfigure}[b]{0.4\linewidth}
		\includegraphics[width=\linewidth]{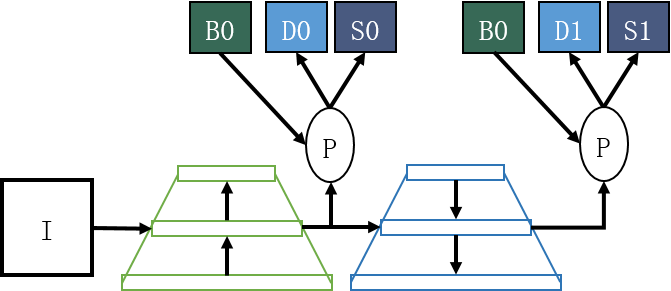}
		\caption{Faster/Mask RCNN w DSFPN}
		\label{fig:faster_mask_w_dsfpn}
	\end{subfigure}
	\hfill
	\begin{subfigure}[b]{0.5\linewidth}
		\includegraphics[width=\linewidth]{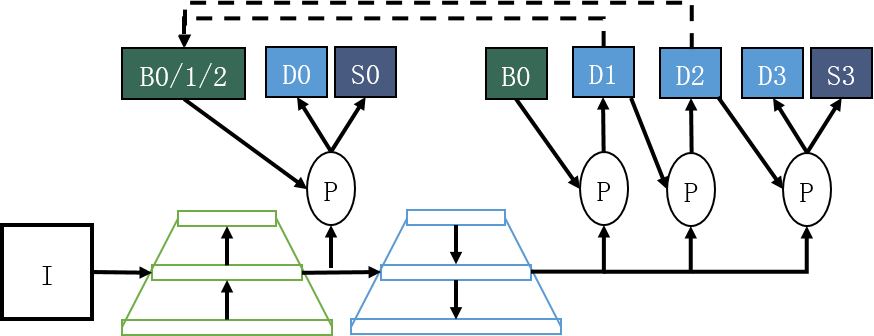}
		\caption{Cascade/Cascade Mask RCNN w DSFPN}
		\label{fig:cascade_mask_w_dsfpn}
	\end{subfigure}
	\caption{The illustration of Faster/Mask/Cascade/CascadeMask RCNN with dually supervised FPN~\cite{lin2017feature} (DSFPN). ``B", ``D", and ``S" indicate proposed \textit{box}, \textit{detection}, and \textit{segmentation} heads, respectively. ``P" is the short for RoI \textit{pooling} operation. ``I" represents the input image. Green and blue pyramid are bottom-up and top-down subnets. The heads mounted on bottom-up pyramid are auxiliary ones.}
	\label{fig:detectors_w_dsfpn}
\end{figure*}

\section{Related work}
\label{relatedwork}
Object detection has been studied for decades, and numerous excellent works have been proposed. In this section, we will review the most relevant of them to highlight the differences of our work.

\noindent
{\bf Two-stage methods.}
The two-stage methods divide the detection into two steps: object proposals and object detection. In general, candidate regions are first yielded; then a detection module classifies these regions and refines the region location. Fast~\cite{girshick2015fast}/Faster RCNN~\cite{ren2015faster} mount a detection head on the top level of deep convolution networks (ConvNets) to take advantage of the rich feature representation for object detection (Fig.~\ref{fig:single_head}). However, this manner neglects the hierarchical characteristics of the deep ConvNets. To utilize this characteristics,  MS-CNN~\cite{cai2016unified} proposes to predict object at multiple levels of deep ConvNets (Fig.~\ref{fig:multi_head}). Although this modification promotes the detectors ability to deal with multi-scales of objects, the low level features lack semantic information for object classification. FPN~\cite{lin2017feature} significantly alleviates this problem by presenting a novel feature pyramid architecture, where low level features are strengthened by a top-down feature pyramid network and object prediction are conducted on each level via lateral connection (Fig.~\ref{fig:u_multi_head}). Further, Mask RCNN~\cite{he2017mask} unifies object detection and segmentation into an end-to-end framework on the basis of FPN~\cite{lin2017feature}. 
More recently, based on FPN~\cite{lin2017feature}, Cascade RCNN and Cascade Mask RCNN~\cite{cai2019cascade,cai2018cascade} propose to yield high quality object detection and segmentation through multi-stage regression.  
Despite superior performance, the feature learning in bottom-up pyramid is not fully explored.

\noindent
{\bf One-stage methods.}
In contrast to two-stage methods, one-stage methods directly perform object detection without proposals, e.g., SSD~\cite{liu2016ssd} and YOLO~\cite{redmon2016you}. Same with Faster RCNN~\cite{ren2015faster}, YOLO~\cite{redmon2016you} predicts objects on the top level of deep ConvNets. Thus, the hierarchy is ignored in detection. 
Similar with MS-CNN~\cite{cai2016unified}, SSD~\cite{liu2016ssd} assigns detection for different scales of objects on different levels of deep ConvNets, which separates the multi-scale objects in the feature space. DSSD~\cite{fu2017dssd} improves SSD~\cite{liu2016ssd} performance by introducing additional context via deconvolution. Further, RefineDet~\cite{zhang2018single} boosts DSSD~\cite{fu2017dssd} using the bottom-up feature pyramid for proposal and the top-down counterpart for detection. However, the major performance gain is from the two-stage box regression but not the bottom-up feature pyramid learning. Moreover, the prediction on bottom-up pyramid is involved in both training and test phase.

\noindent
{\bf Supervision in deep ConvNets.}
Usually, the feature learning in deep ConvNets is guided by a supervision signal from the output layer, e.g., VGG~\cite{simonyan2014very}, ResNet~\cite{he2016deep} (Fig.~\ref{fig:single_head}), and UNet~\cite{ronneberger2015u} (Fig.~\ref{fig:u_single_head}). This results in the gradient vanishing with the increasing of depth of layers, such that the network is hard to effectively train. While ResNet~\cite{he2016deep} alleviates the the problem via skip connection, it still remains a challenge to train deep ConvNets. DSN~\cite{lee2015deeply} solves this problem in general classification task by proposing a deeply-supervised training architecture, where auxiliary classifiers are introduced to hidden layers (Fig.~\ref{fig:multi_head}). PSPNet~\cite{zhao2017pyramid} demonstrates that adding an extra loss at the relatively high level layer of deep ConvNets can benefit the performance improvement in scene parsing.
The similar idea is also exploited in HED~\cite{xie2015holistically} for edge detection, but the prediction heads are attached on each level of deep ConvNets. Additionally, in HED~\cite{xie2015holistically} all the heads are needed in inference phase.

\noindent
{\bf Difference from previous works.}  
To highlight the differences of proposed dually supervised FPN (DSFPN) from previous related works, we summarize the typical network architecture for object classification, detection and segmentation in Fig.~\ref{fig:typical_architecture}. For better illustration, the skip connection is omitted in the backbone network. 
By comparing with Fig.~\ref{fig:single_head} and \ref{fig:u_single_head}, where the supervision signal is simply at the final output layer, DSFPN has multiple supervision signals on every layer of the network. Although in Fig.~\ref{fig:multi_head} the supervision signals are also on each layer, this structure is not suited for object detection or segmentation, due to the shortage of representation ability in lower layers. Note that SSD~\cite{liu2016ssd} adopts a similar architecture to Fig.~\ref{fig:multi_head} but uses relatively higher layers. In contrast to Fig.~\ref{fig:u_multi_head}, DSFPN (Fig.~\ref{fig:dsfpn}) adds auxiliary supervision signals on inner layers, such that the network can be trained more effectively.    

\section{Method}
\label{method}

In this section we elaborate the proposed \textit{dually supervised FPN} (DSFPN) in details and present its formulation.
\subsection{Overview}
Generally speaking, DSFPN is proposed to deal with the gradient degradation in feature pyramid network for object detection or segmentation at region level during training phase. To this end, when proposals are given, auxiliary prediction heads (\ie, object detection or segmentation heads) are attached to the bottom-up subnet (Fig.~\ref{fig:detectors_w_dsfpn}), such that supervision signal can immediately direct the feature learning in it. In the test phase, these auxiliary heads are discarded so that no extra computation is needed in model inference.
Formally, given a set of $N$ proposal boxes $B=\{b_i| i=1, 2, ..., N\}$, they are projected to a level of feature layer according to the box size. Here we use the $k$-th level as an example, the boxes on the $k$-th level are $B_k$. The feature maps of the $k$-th level are denoted as $f^k$. Thus, the detection and segmentation heads on top-down pyramid can be formulated as $D_{td}^k(f_{td}^k,B_k)$ and $S_{td}^k(f_{td}^k,B_k)$, respectively. Accordingly, the prediction heads on bottom-up pyramid are $D_{bu}^k(f_{bu}^k,B_k)$ and $S_{bu}^k(f_{bu}^k,B_k)$. 
In the training phase, the predictions over the both pyramid subnets are $D=\{D_{td}^k, D_{bu}^k|k=1, 2, ..., M\}$ and $S=\{S_{td}^k, S_{bu}^k|k=1, 2, ..., M\}$, where $M$ is the number of levels of pyramid. Here, `td' and `bu' indicate top-down and bottom-up, respectively. In the inference phase, the predictions are $D=\{D_{td}^k|k=1, 2, ..., M\}$ and $S=\{S_{td}^k|k=1, 2, ..., M\}$, since $D_{bu}^k$ and $S_{bu}^k$ are removed. 
The given proposal boxes are not necessary from proposal stage (\eg, RPN~\cite{ren2015faster}), they can also come from latter stages. The effect of choosing the boxes from different stages will be discussed in Experiment. Since the object or segmentation heads over different levels share same weights, the $k$ in each head notation is dropped for simplicity in subsequent sections.

\subsection{Dual supervision}
The proposed DSFPN is general to two- and multi-stage detectors, we describe the concrete architecture one by one.

\begin{figure}[t]
	\begin{center}
	\includegraphics[width=1\linewidth]{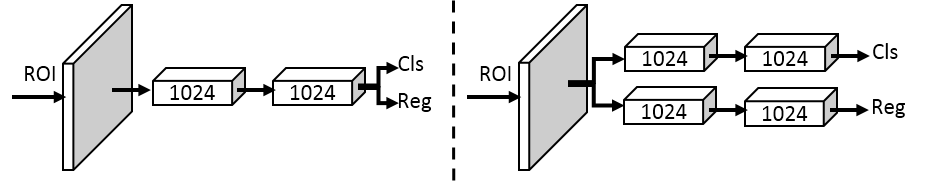}
	\end{center}
	\caption{The structure of detection head in FPN~\cite{lin2017feature} (left) and the proposed DSFPN (right).}
	\label{fig:cls_reg_decouple}
\end{figure}

\noindent
{\bf Two-stage detector.}
In two-stage detectors, we use two representative methods, Faster RCNN~\cite{ren2015faster} and Mask RCNN~\cite{he2017mask}, as examples to illustrate the process of incorporating  dual supervision (DS). As shown in Fig.~\ref{fig:faster_mask_w_dsfpn}, the proposal boxes, $B0$, are first given and projected to each level of pyramid. In Faster/Mask RCNN~\cite{ren2015faster,he2017mask}, $B0$ is yielded by RPN~\cite{ren2015faster}, which is not displayed in Fig.~\ref{fig:faster_mask_w_dsfpn}. The detection and segmentation over both top-down and bottom-up pyramids are $D1$, $S1$, $D0$, and $S0$, which correspond to $D_{td}$, $S_{td}$, $D_{bu}$, and $S_{bu}$, respectively. The final loss of the detector can be written as 
\begin{equation} 
\label{eq1}
\begin{aligned}
\mathcal{L}_{final}=\alpha_1\mathcal{L}(D_{gt},D0)+\alpha_2\mathcal{L}(D_{gt},D1)\\
+\alpha_3\mathcal{L}(S_{gt},S0)+\alpha_4\mathcal{L}(S_{gt},S1)
\end{aligned}
\end{equation}
where the $D_{gt}$ and $S_{gt}$ represent ground truth of detection and segmentation; $\mathcal{L}(D,\cdot)$ and $\mathcal{L}(S,\cdot)$ are the detection and segmentation loss functions; $\alpha_1, \alpha_2, \alpha_3, \alpha_4$ are the weights for each loss. The specific formulas of the two functions are referred to~\cite{he2017mask}. The $\mathcal{L}(S,\cdot)$ can be omitted when segmentation ground truth is not available.

\noindent
{\bf Multi-stage detector.}
Recently proposed Cascade and Cascade Mask RCNN~\cite{cai2018cascade,cai2019cascade} are used to demonstrate how to integrate DS into multi-stage detectors (Fig.~\ref{fig:cascade_mask_w_dsfpn}).
The proposed boxes can be produced from RPN~\cite{ren2015faster}, detection head at first stage ($D1$) or second stage ($D2$), which are denoted as $B0$, $B1$, and $B2$, respectively. On top-down pyramid, the proposed boxes can only be $B0$.
However, on bottom-up pyramid, the boxes can be one of the $B0, B1$, and $B2$.
Similar to two-stage detectors, the predictions over bottom-up pyramid are $D0$ and $S0$. However, the predictions over top-down pyramid are separated into three sets: $D1$, $D2$, and $D3$. As stated in Cascade Mask RCNN~\cite{cai2019cascade}, the segmentation head is simply attached at stage three (denoted as $S3$) for the trade-off between accuracy and computation complexity.
Thus, the final loss of the detectors is written as
\begin{equation} 
\label{eq2}
\begin{aligned}
\mathcal{L}_{final}=&\alpha_1\mathcal{L}(D_{gt},D0)+\alpha_2\mathcal{L}(S_{gt},S0)\\
 &+\alpha_3\mathcal{L}(S3_{gt},S3)+
\sum_{i=1}^{T}\alpha_{si}\mathcal{L}(Di_{gt},Di)
\end{aligned}
\end{equation}
where $i$ is the index of the prediction stage on the top-down pyramid; $T$ is the total number of stages;  $S3$ and $S3_{gt}$ are the segmentation predictions at stage three and its corresponding ground truth; $Di$ and $Di_{gt}$ are detections and  ground truth at stage $i$; $\mathcal{L}(D,\cdot)$ and $\mathcal{L}(S,\cdot)$ are the detection and segmentation loss functions; $\alpha_1, \alpha_2, \alpha_3, \alpha_{si}$ are weights of each prediction head. Segmentation related terms only exist in Cascade Mask~RCNN~\cite{cai2019cascade}.
\vspace{-0cm}
\begin{table*}[t]
	\centering
	\footnotesize
	\caption{The \textbf{ablation study} in representative \textbf{two-stage detectors}, \ie, Faster RCNN and Mask RCNN. $1\times$ and $2\times$ indicate 12 epochs and 24 epochs of training schedule, respectively. $\boldsymbol{\star}$ indicates multi-scale testing.}
	\begin{tabular}{c|c|c|c|cccccc|cccccc}
		\hline
		\hline
		Method&backbone&DS&DC&AP$_{bbox}$&AP$_{50}$&AP$_{75}$&AP$_S$&AP$_M$&AP$_L$&AP$_{mask}$&AP$_{50}$&AP$_{75}$\\
		\hline
	    \multirow{4}{*}{Faster RCNN$_{1\times}$}&\multirow{4}{*}{ResNet50}&\xmark&\xmark&36.7&58.4&39.6&21.1&39.9&48.1&-&-&-\\
	    
	    &&\cmark&\xmark&37.7&59.0&40.6&20.6&40.6&40.6&-&-&-\\
	    
	    &&\xmark&\cmark&37.5&58.8&40.8&21.1&40.7&49.2&-&-&-\\
	    
	    &&\cmark&\cmark&\textbf{38.0}&\textbf{59.3}&\textbf{41.3}&\textbf{21.3}&\textbf{40.9}&\textbf{49.7}&-&-&-\\
	    \hline
	    \multirow{2}{*}{Faster RCNN $_{2\times}$}&\multirow{2}{*}{ResNet50}&\xmark&\xmark&37.9&59.3&41.1&21.5&41.1&49.9&-&-&-\\
	    &&\cmark&\cmark&\textbf{39.2}&\textbf{60.1}&\textbf{42.7}&\textbf{22.2}&\textbf{42.3}&\textbf{52.1}&-&-&-\\
	    \hline
	    \multirow{2}{*}{Faster RCNN $_{1\times}$}&\multirow{2}{*}{ResNet101}&\xmark&\xmark&39.4&61.2&43.4&22.6&42.9&51.4&-&-&-\\
	    &&\cmark&\cmark&\textbf{40.0}&61.2&43.4&22.6&\textbf{43.3}&\textbf{53.3}&-&-&-\\
        \hline
	    \multirow{2}{*}{Faster RCNN $_{2\times}$}&\multirow{2}{*}{ResNet101}&\xmark&\xmark&39.8&61.3&43.3&22.9&43.3&52.6&-&-&-\\
	    &&\cmark&\cmark&\textbf{41.1}&\textbf{62.2}&\textbf{44.3}&\textbf{23.2}&\textbf{44.4}&\textbf{54.8}&-&-&-\\
	    \hline
	    \multirow{2}{*}{Faster RCNN$_{1\times}$} &\multirow{2}{*}{ResNeXt101}&\xmark&\xmark&41.5&63.8&44.9&24.8&45.4&53.5&-&-&-\\
	    
	    &&\cmark&\cmark&\textbf{42.5}&63.8&\textbf{46.3}&\textbf{25.2}&\textbf{46.4}&\textbf{55.7}&-&-&-\\
	    \hline
	    \multirow{2}{*}{Faster RCNN$_{2\times}$} &\multirow{2}{*}{ResNeXt101}&\xmark&\xmark&40.8&62.5&44.4&23.4&44.3&53.8&-&-&-\\
	    
	    &&\cmark&\cmark&\textbf{42.6}&\textbf{63.5}&\textbf{46.4}&\textbf{24.7}&\textbf{46.5}&\textbf{56.0}&-&-&-\\
	    \hline
	    \multirow{2}{*}{Faster RCNN$_{2\times}$$\boldsymbol{\star}$} &\multirow{2}{*}{ResNeXt101}&\xmark&\xmark&44.0&65.0&47.9&28.7&46.9&57.4&-&-&-\\
	    
	    &&\cmark&\cmark&\textbf{45.4}&\textbf{65.4}&\textbf{49.2}&\textbf{29.4}&\textbf{48.5}&\textbf{58.9}&-&-&-\\
	    
	    \hline
	    \hline
	    \multirow{2}{*}{Mask RCNN$_{1\times}$} &\multirow{2}{*}{ResNet50}&\xmark&\xmark&37.7&59.2&40.9&21.4&40.8&49.8&33.9&55.8&35.8\\
	    
	     &&\cmark&\cmark&\textbf{38.8}&\textbf{59.6}&\textbf{42.3}&\textbf{21.5}&\textbf{41.6}&\textbf{51.7}&\textbf{34.6}&\textbf{56.0}&\textbf{37.0}\\
	    \hline
	    \multirow{2}{*}{Mask RCNN $_{2\times}$}&\multirow{2}{*}{ResNet50}&\xmark&\xmark&38.6&59.8&42.1&22.2&41.5&50.8&34.5&56.5&36.3\\
	    
	    &&\cmark&\cmark&\textbf{40.0}&\textbf{60.9}&\textbf{43.1}&\textbf{22.4}&\textbf{42.6}&\textbf{54.1}&\textbf{35.6}&\textbf{57.6}&\textbf{37.7}\\
	    \hline
	    \multirow{2}{*}{Mask RCNN$_{1\times}$} &\multirow{2}{*}{ResNet101}&\xmark&\xmark&40.0&61.8&43.7&22.5&43.4&52.7&35.9&58.3&38.0\\
	    
	    &&\cmark&\cmark&\textbf{40.7}&\textbf{61.5}&\textbf{44.6}&\textbf{22.5}&\textbf{43.8}&\textbf{54.5}&\textbf{36.3}&\textbf{58.3}&\textbf{38.5}\\
	    \hline
	    \multirow{2}{*}{Mask RCNN$_{2\times}$} &\multirow{2}{*}{ResNet101}&\xmark&\xmark&40.9&61.9&44.8&23.5&44.2&53.9&36.4&58.3&38.7\\
	    &&\cmark&\cmark&\textbf{42.1}&\textbf{62.6}&\textbf{45.8}&\textbf{23.9}&\textbf{45.3}&\textbf{56.1}&\textbf{37.1}&\textbf{59.2}&\textbf{39.5}\\
	    \hline
	    \multirow{2}{*}{Mask RCNN$_{1\times}$} &\multirow{2}{*}{ResNeXt101}&\xmark&\xmark&42.4&64.3&46.5&25.5&46.7&54.6&37.5&60.6&39.9\\

	    &&\cmark&\cmark&\textbf{42.8}&63.5&\textbf{46.7}&\textbf{25.0}&\textbf{46.5}&\textbf{56.3}&\textbf{37.6}&60.1&39.9\\
	    \hline
	    \multirow{2}{*}{Mask RCNN$_{2\times}$} &\multirow{2}{*}{ResNeXt101}&\xmark&\xmark&42.3&63.6&45.9&25.4&45.9&55.7&37.2&59.9&39.4\\

	    &&\cmark&\cmark&\textbf{43.3}&\textbf{63.9}&\textbf{47.1}&\textbf{25.9}&\textbf{46.8}&\textbf{57.0}&\textbf{37.9}&\textbf{60.4}&\textbf{40.4}\\
	    \hline
	\end{tabular}
	\label{table:results_faster_mask_rcnn}
\end{table*}

\subsection{Decoupled head}
For detection task, the architecture of the \textit{cls}-\textit{reg} decoupled head is shown in Fig.~\ref{fig:cls_reg_decouple} right.  Comparing with the original detection head structure (Fig.~\ref{fig:cls_reg_decouple} left), the decoupled head separates the classification and regression tasks in hidden feature space, which is achieved by taking apart the shared two hidden layers with 1,024 nodes. Formally, the detection head is written as 
\begin{equation} 
\label{eq3}
\begin{aligned}
D(B,f)&=C(B,f)+R(B,f)\\
C(B,f)&=W_{c}^2\cdot W_{c}^1\cdot f^B\\
R(B,f)&=W_{r}^2\cdot W_{r}^1\cdot f^B
\end{aligned}
\end{equation}
 where $C$ and $R$ symbolize classification and regression, respectively; $W_c$ and $W_r$ represent the weights of hidden layers of classification and regression, respectively; `1' and `2' denote indices of hidden layers; $f^B$ is the feature in box regions.
This simple operation increases the capacity of detection head to address two inconsistent tasks. Moreover, the decoupling disentangles the propagation of supervision signal until the RoI pooling.

\begin{table*}[t]
\centering
	\footnotesize
	\caption{The \textbf{ablation study} in representative \textbf{multi-stage detectors}, \ie, Cascade RCNN and Cascade Mask RCNN. $1\times$ and $2\times$ indicate 12 epochs and 24 epochs of training schedule, respectively.}
	\begin{tabular}{c|c|c|c|cccccc|cccccc}
		\hline
		\hline
		Method&backbone&DS&DC&AP$_{bbox}$&AP$_{50}$&AP$_{75}$&AP$_S$&AP$_M$&AP$_L$&AP$_{mask}$&AP$_{50}$&AP$_{75}$\\
		\hline
	    \multirow{2}{*}{Cascade RCNN $_{1\times}$}&\multirow{2}{*}{ResNet50}&\xmark&\xmark&40.9&59.0&44.6&22.5&43.6&55.3&-&-&-\\
	    &&\cmark&\cmark&\textbf{41.3}&\textbf{59.6}&\textbf{44.9}&\textbf{22.6}&\textbf{44.1}&\textbf{55.8}&-&-&-\\
	    \hline
	    \multirow{2}{*}{Cascade RCNN $_{2\times}$}&\multirow{2}{*}{ResNet50}&\xmark&\xmark&41.3&59.7&44.9&23.8&44.0&55.3&-&-&-\\
	    
	    &&\cmark&\cmark&\textbf{42.1}&\textbf{60.5}&\textbf{45.6}&\textbf{23.5}&\textbf{44.8}&\textbf{56.9}&-&-&-\\
	    \hline
	    \multirow{2}{*}{Cascade RCNN$_{1\times}$} &\multirow{2}{*}{ResNet101}&\xmark&\xmark&42.8&61.4&46.8&24.1&45.8&57.4&-&-&-\\
	    &&\cmark&\cmark&\textbf{43.1}&\textbf{61.7}&\textbf{47.0}&\textbf{24.4}&\textbf{46.5}&\textbf{57.8}&-&-&-\\
	    \hline
	    \multirow{2}{*}{Cascade RCNN $_{2\times}$}&\multirow{2}{*}{ResNet101}&\xmark&\xmark&43.0&61.4&46.9&23.9&46.1&58.0&-&-&-\\
	    
        &&\cmark&\cmark&\textbf{43.7}&\textbf{62.3}&\textbf{47.5}&\textbf{24.7}&\textbf{47.0}&\textbf{58.8}&-&-&-\\
	    \hline
	    \hline
	    \multirow{2}{*}{CascadeMask RCNN $_{1\times}$}&\multirow{2}{*}{ResNet50}&\xmark&\xmark&41.3&59.6&44.9&23.1&44.2&55.4&35.4&56.2&37.8\\
	    &&\cmark&\cmark&\textbf{42.0}&\textbf{60.1}&\textbf{45.8}&\textbf{23.4}&\textbf{45.1}&\textbf{56.6}&\textbf{35.9}&\textbf{56.7}&\textbf{38.4}\\
	    \hline

	    \multirow{2}{*}{CascadeMask RCNN $_{1\times}$}&\multirow{2}{*}{ResNet101}&\xmark&\xmark&43.3&61.7&47.2&24.2&46.3&58.2&37.1&58.6&39.8\\
	    &&\cmark&\cmark&\textbf{43.6}&\textbf{61.9}&\textbf{47.4}&\textbf{24.6}&\textbf{46.8}&\textbf{58.5}&\textbf{37.3}&58.5&\textbf{40.0}\\
	    \hline
	    \multirow{2}{*}{CascadeMask RCNN $_{2\times}$}&\multirow{2}{*}{ResNet101}&\xmark&\xmark&43.5&62.1&47.2&24.3&46.9&58.2&37.2&58.7&39.5\\
	    
	    &&\cmark&\cmark&\textbf{44.5}&\textbf{63.0}&\textbf{48.4}&\textbf{24.9}&\textbf{47.8}&\textbf{60.0}&\textbf{38.1}&\textbf{59.6}&\textbf{41.0}\\
	    \hline
	\end{tabular}
	\label{table:results_cascade_mask_rcnn}
\end{table*}

\section{Experiments}
\label{experiment}
\begin{figure}[t]
	\begin{center}
	\includegraphics[width=0.83\linewidth,height=.55\linewidth]{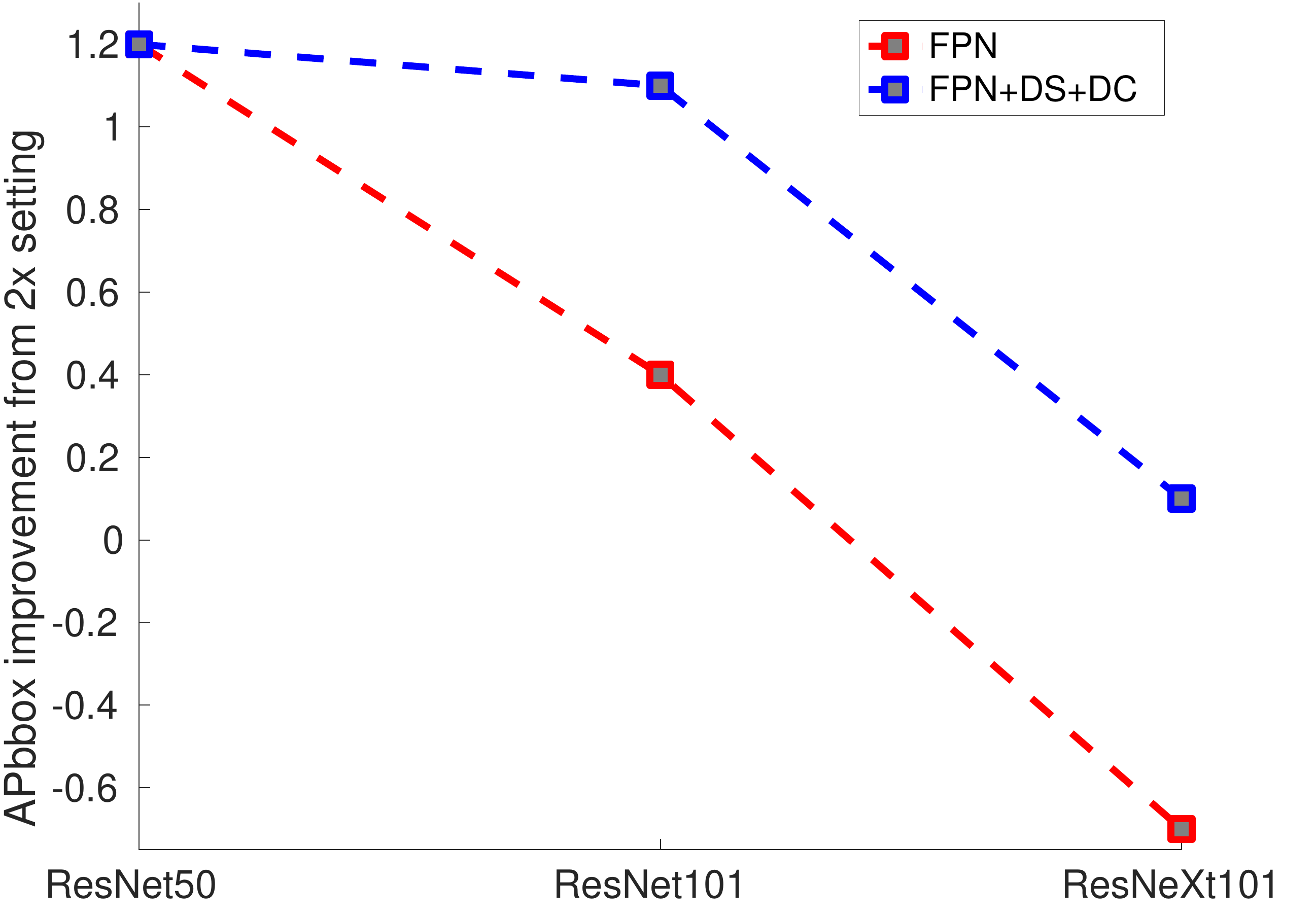}
	\end{center}
	\vspace{-2mm}
	\caption{The effect of DS+DC on performance change from $2\times$ setting. The detector is Faster RCNN.}
	\label{fig:APbbox_improvement}
\end{figure}
\subsection{Implementation details}
We perform experiments on the MS COCO dataset~\cite{lin2014microsoft} with 80 categories. The union of 80k train images and 35k images of val images (train2017) is used as train dataset. The 5k of val (val2017) is utilized to report results of ablation study. The final results are reported on the test-dev dataset for fair comparison with state-of-the-art detectors. 
We implement our dually supervised FPN (DSFPN) based on the detetron\footnote{https://github.com/facebookresearch/Detectron} library and Caffe2\footnote{https://caffe2.ai/} framework. By convention, we set the short side of an input image to 800 pixels and the other side is less than $1,333$ pixels. The number of anchors and proposals are set as default. The RoIAlign in~\cite{he2017mask} is adopted in all experiments. The backbone network of all detectors includes FPN~\cite{lin2017feature}. For Faster/Mssk/Cascade RCNN based on ResNet50/ResNet101~\cite{he2016deep}, synchronized SGD is used to train the model on 8 P100 GPUs, each mini-batch involves 2 images per GPU. The other detectors are trained with 1 image per GPU, due to the limited GPU memory. No data augmentation is used except for horizontal flipping. Both $1\times$ (\ie, 12 epochs) and $2\times$ (\ie, 24 epochs) training schedules are utilized in experiments. The weights of each item of final loss are set to 1. In inference, all experiments are conducted on a single image scale ($800\times1,333$ pixels) without bells and whistles, unless otherwise noted.

\subsection{Ablation study}

We validate the effects of dual supervision (DS) and decouple (DC) operation on  Faster/Mask RCNN~\cite{ren2015faster,he2017mask} and Cascade/CascadeMask RCNN~\cite{cai2018cascade,cai2019cascade}. The generalizability of the proposed DSFPN is evaluated by comparing experimental results on the val2017 and test-dev datasets.  The inference time is measured on a 1080Ti GPU with input size $800\times1,333$ pixels.

\noindent
{\bf Effects of DS and DC on Faster/Mask RCNN.}
From Table~\ref{table:results_faster_mask_rcnn}, we note that DS and DC improve the baseline, Faster RCNN$_{1\times}$ with ResNet50~\cite{he2016deep}, by 1 and 0.8 points in AP$_{bbox}$, respectively. Combining them together, the improvement in AP$_{bbox}$ reaches 1.3 points. By conducting experiments over various baseline, backbone, and training schedule, we find that the  increment is consistent in terms of AP$_{bbox}$ over different IoUs and scales. An interesting observation is that in baselines, the deeper the backbone, the less the effect of doubling training schedule ($2\times$) (Fig.~\ref{fig:APbbox_improvement}). For example, when backbone is ResNet50 and detector is Faster RCNN, the $2\times$ setting increases the AP$_{bbox}$ by $1.2$ points; when backbone is ResNet101, the incremental margin declines to $0.4$ points; when backbone is ResNeXt101, the incremental magnitude even becomes a negative value ($-0.7$ points).  
This is because the deeper the network, the harder the training.
In contrast, when the baselines are equipped with DS and DC, the $2\times$ setting consistently improves AP$_{bbox}$ over all backbones.
This can be attributed to the promotion of model training via enhanced supervision signal.  In addition to detection, Mask RCNN with DS and DC also outperforms all the ones without the two operations in terms of AP$_{mask}$, and the effect on $2\times$ is apparent as well.

\noindent
{\bf Effects of DS and DC on Cascade/CascadeMask RCNN.}
As shown in Table~\ref{table:results_cascade_mask_rcnn}, promising improvement of AP$_{bbox}$ is gained in all baseline detectors by adding DS and DC. Similar to the two-stage detectors, when DS and DC are involved in model, detectors can benefit more from the double training schedule ($2\times$), especially for deeper backbones. Specifically, when backbone network is ResNet101 and detector is Cascade RCNN, $2\times$ helps the model with DS and DC to obtain a higher improvement margin than that without DS and DC ($43.7$ vs $43.1$ and $43.0$ vs $42.8$).  By comparing the performance in terms of AP$_{50}$ and AP$_{75}$ in detection task, we see that adding DS and DC can achieve larger increment in AP$_{50}$ than that in AP$_{75}$. This can be attributed to the multi-stage regression yielding high-quality box, such that little room remains to explore in this direction. In the segmentation task, adding DP and DC can still lead to performance improvement, even if the backbone is ResNet101.  
\begin{table}[t]
	\centering
	\footnotesize
	\caption{The effect of boxes in the bottom-up pyramid. }
	\begin{tabular}{c|c|c|ccc}
		\hline
		\hline
		&backbone&stage&AP& AP$_{50}$&AP$_{75}$\\
		\hline
	    Cascade RCNN $1\times$ &ResNet50&0&41.3&59.6&44.9\\
	    \hline
	    Cascade RCNN $1\times$ &ResNet50&1&41.0&59.2&44.3\\
	    \hline
	    Cascade RCNN $1\times$ &ResNet50&2&40.9&59.0&44.1\\
	    \hline
	\end{tabular}
	\label{table:effect_box_stage012}
\end{table}

	    
	    
\begin{table*}[t]
\centering
\caption{Generalizability of the proposed DSFPN.}
\resizebox{1\textwidth}{!}{
\begin{tabular}{c|c|c|c|cccccc|cccccc}
\hline
\hline
\multirow{2}{*}{method} & \multirow{2}{*}{backbone} & \multirow{2}{*}{DSFPN} & \multirow{2}{*}{test speed} & \multicolumn{6}{c|}{val2017} & \multicolumn{6}{c}{test-dev} \\ \cline{5-16} 
 &  &  &  & $AP_{bbox}$ & $AP_{50}$ & $AP_{75}$ & $AP_{mask}$ & $AP_{50}$ & $AP_{75}$ & $AP_{bbox}$ & $AP_{50}$ &$ AP_{75}$ &$AP_{mask}$  & $AP_{50}$ & $AP_{75}$ \\ 
\hline
\multirow{2}{*}{FasterRCNN$_{2\times}$} & \multirow{2}{*}{ResNet50} &  \xmark&  0.074s&37.9  &59.3  &41.1  &-  &-  &-  &37.9  &59.6  &41.0  &-  &-  &-  \\

 & & \cmark &0.075s& 39.2 & 60.1 & 42.7 &-  &-  &-  &39.3  &60.5  &42.8  &-  &-  &- \\
 \hline
 \multirow{2}{*}{FasterRCNN$_{2\times}$} & \multirow{2}{*}{ResNet101} &  \xmark& 0.097s &39.8  & 61.3 &43.3  & - & - & - & 40.0 & 61.7 & 43.5 & - & - & - \\

 &  & \cmark &0.098s  & 41.1 & 62.2&44.3&-  &-  &-  &41.3  &62.5  &44.5&-  &-  &- \\
 \hline
 \multirow{2}{*}{FasterRCNN$_{2\times}$} & \multirow{2}{*}{ResNeXt101} &  \xmark&  0.202s&40.8  &62.5  &44.4  &-  & - & - &41.1  &63.1  &44.6  & - & - & - \\

 &  &\cmark  &  0.203s&42.6  &63.5  &46.4  & - &-  &-  &43.0  &64.4  &46.7  &-  &-  &- \\
 \hline
 \multirow{2}{*}{MaskRCNN$_{2\times}$} & \multirow{2}{*}{ResNet50} &  \xmark&  0.082s&38.6  &59.8  &42.1  &34.5  &56.5  &36.3  &39.0  &60.5  &42.6  &34.9  &57.2  &36.9  \\

 &  & \cmark &  0.084s&40.0  &60.9  &43.1  &35.6  &57.6  &37.7  &40.2  &61.1  &43.7  &35.7  &57.8  &37.8 \\
 \hline
 \multirow{2}{*}{MaskRCNN$_{2\times}$} & \multirow{2}{*}{ResNet101} &  \xmark&  0.104s& 40.9 &61.9  & 44.8 & 36.4 & 58.5 & 38.7 & 41.0 &62.4  &44.8  &36.4  &59.1  &38.6  \\

 &  & \cmark & 0.105s &42.1  &62.6  & 45.8 &37.1  & 59.2 &39.5  &42.3  &63.0  &45.9  &37.3  &59.8  &39.6 \\
 \hline
 \multirow{2}{*}{MaskRCNN$_{2\times}$} & \multirow{2}{*}{ResNeXt101} &  \xmark&  0.210s&42.3  &63.6  &45.9  &37.2 &59.9  &39.4  &42.6  &64.1  &46.5  &37.6  &60.8  &39.9  \\

 &  & \cmark & 0.213s &43.3  &63.9  &47.1  &37.9  &60.4  &40.4  &43.8  &64.7  &47.9  &38.4  &61.5  &40.9 \\
 \hline
 \hline
 \multirow{2}{*}{CascadeRCNN$_{2\times}$} & \multirow{2}{*}{ResNet50} &  \xmark& 0.101s &41.3  &59.7  &44.9 &-  &-  &-  &41.5  &59.9  &45.2  &-  &-  &-  \\

 &  &  \cmark&  0.115s&42.1  &60.8  &45.6  & - & - &-  &42.1  &60.8  &45.7  &-  &-  &- \\
 \hline
 \multirow{2}{*}{CascadeRCNN$_{2\times}$} & \multirow{2}{*}{ResNet101} &  \xmark&  0.123s&  43.0& 61.4 & 46.9 &-  &-  &-  &43.0  &61.3  &47.0  &-  &-  &-  \\
 &  & \cmark &0.137s  &43.7  &62.3  &47.5  & - & - & - & 43.7 & 62.4 & 47.4 & - &-  &- \\
 \hline

 \multirow{2}{*}{CascadeMaskRCNN$_{2\times}$} & \multirow{2}{*}{ResNet101} & \xmark &0.131s&43.5  &62.1  &47.2  &37.2  &58.7  &39.5  &43.7  &62.3  &47.6  &37.4  &59.4  & 40.0 \\

 &  &\cmark  &0.147s  &44.5  &63.0  &48.4  &38.1  &59.6  &41.0  &44.5  &63.0  &48.3&38.0  &60.0  &40.8 \\
 \hline
\end{tabular}
}
\label{table:generalizability}
\end{table*}

\begin{figure*}[t]
	\begin{center}
	\includegraphics[width=1\linewidth]{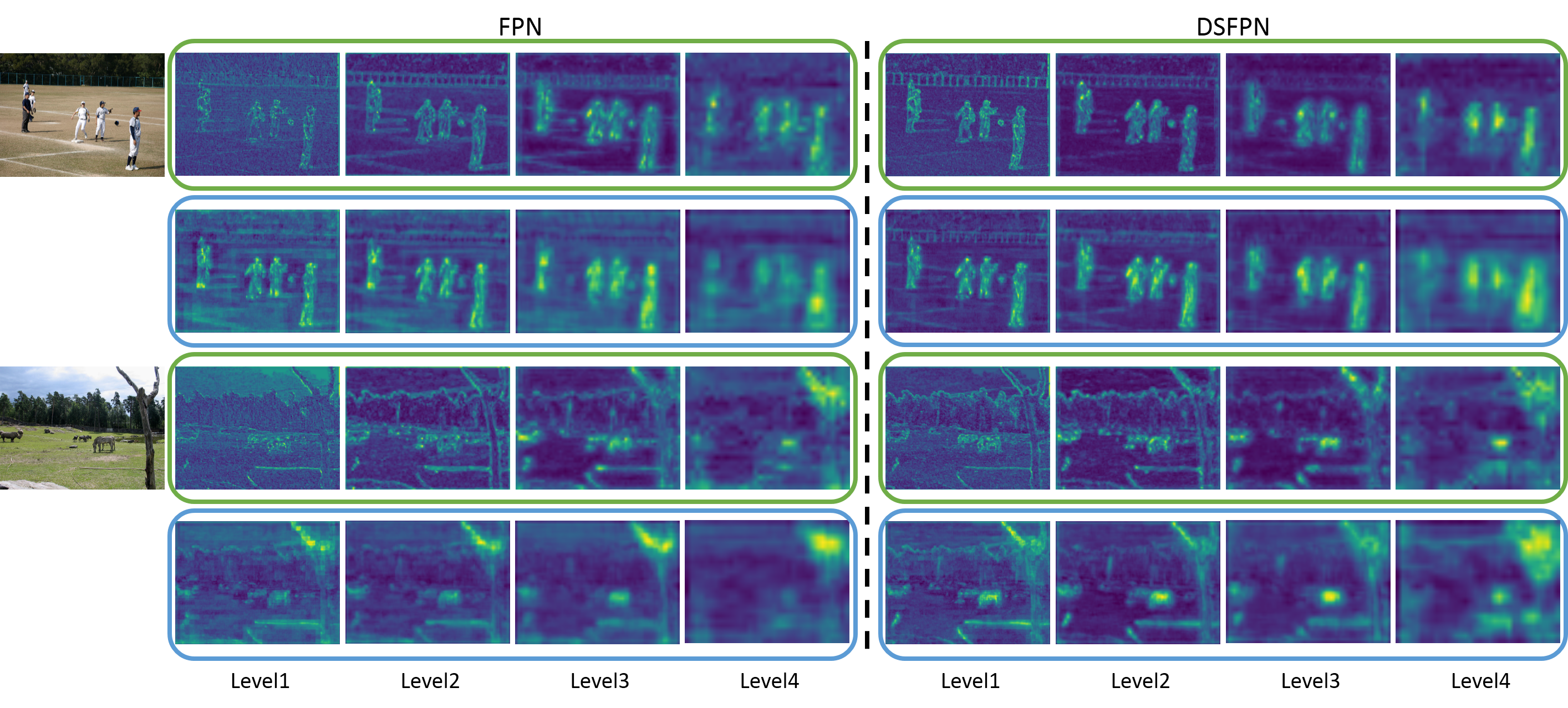}
	\end{center}
	\vspace{-6mm}
	\caption{The feature map of each layers in FPN~\cite{lin2017feature} (left) and DSFPN (right). The green and blue boxes indicate that the feature maps enclosed are from bottom-up and top-down pyramids, respectively.}
	\label{fig:det_feat}
	\vspace{-3mm}
\end{figure*}

\begin{table*}[t]
	\centering
	\footnotesize
	\caption{Comparison with state-of-the-art methods on the COCO~\cite{lin2014microsoft} test-dev dataset.}  
	\begin{tabular}{r|c|ccccccccc}
		\hline
		\hline
		method&backbone&AP&AP$_{50}$&AP$_{75}$&AP$_S$&AP$_M$&AP$_L$\\
	    \hline
	    YOLOv2~\cite{redmon2017yolo9000}&DarkNet-19&21.6&44.0&19.2&5.0&22.4&35.5\\
	    SSD-513~\cite{liu2016ssd} &ResNet101&31.2&50.4&33.3&10.2&34.5&49.8\\
	    DSSD-513~\cite{fu2017dssd} &ResNet101&33.2&53.3&35.2&13.0&35.4&51.1\\
	    RefineDet512~\cite{zhang2018single} &ResNet101&36.4&57.5&39.5&16.6&39.9&51.4\\
	    RetinaNet800~\cite{lin2017focal} &ResNet101&39.1&59.1&42.3&21.8&42.7&50.2\\
	    CornerNet~\cite{law2018cornernet} &Hourglass-104&40.5&56.5&43.1&19.4&42.7&53.9\\
	    FASF800~\cite{zhu2019feature}&ResNeXt101&42.9&63.8&46.3&26.6&46.2&52.7\\
	    \hline
	    Faster RCNN+++~\cite{he2016deep}&ResNet101&34.9&55.7&37.4&15.6&38.7&50.9\\
	    Faster RCNN w FPN~ \cite{lin2017feature}&ResNet101&36.2&59.1&39.0&18.2&39.0&48.2\\
	    Mask RCNN~\cite{he2017mask} &ResNeXt101&39.8&59.8&-&22.1&43.2&51.2\\
	    Grid RCNN~\cite{lu2019grid} w FPN&ResNet101&41.5&60.9&44.5&23.3&44.9&53.1\\
	    Grid RCNN~\cite{lu2019grid}w FPN&ResNeXt101&43.2&63.0&46.6&25.1&46.5&55.2\\
	    Libra RCNN~\cite{pang2019libra} w FPN&ResNet101&41.1&62.1&44.7&23.4&43.7&52.5\\
	    Libra RCNN~\cite{pang2019libra} w FPN&ResNeXt101&43.0&64.0&47.0&25.3&45.6&54.6\\
	    Cascade RCNN~\cite{cai2018cascade} w FPN&ResNet101&42.8&62.1&46.3&23.7&45.5&55.2\\
	    TridentNet~\cite{li2019scale}&ResNet101&42.7&63.6&46.5&23.9&46.6&56.6\\
	    Cascade RCNN w Rank-NMS~\cite{Tan_2019_ICCV}&ResNet101&43.2&61.8&47.0&24.6&46.2&55.4\\
	    \hline
	    Faster RCNN w DSFPN (ours) &ResNeXt101&43.0&64.4&46.7&25.3&46.1&54.7\\
	    Mask RCNN w DSFPN (ours) &ResNet101&42.3&63.0&45.9&23.4&44.9&54.6\\
	    Mask RCNN w DSFPN (ours) &ResNeXt101&43.8&\textbf{64.7}&47.9&\textbf{25.5}&46.8&55.7\\
	    Cascade RCNN w DSFPN (ours) &ResNet101&43.7&62.4&47.4&24.3&45.9&56.8\\
	    Cascade Mask RCNN w DSFPN (ours) &ResNet101&\textbf{44.5}&63.0&\textbf{48.3}&24.3&\textbf{47.2}&\textbf{58.5}\\
	    \hline
	    
	\end{tabular}
	\label{table:compare_sota}
\end{table*}
\vspace{-0.5cm}
\noindent
{\bf Boxes generated from stage 0, 1, and 2.}
The experimental results with boxes from different stages are listed in Table~\ref{table:effect_box_stage012}. We note that (1) using boxes from stage 0 (\ie, RPN~\cite{ren2015faster}) achieves best performance; and (2) the higher the stage, the lower the performance. It suggests that when the boxes are coarse (\ie, boxes from stage 0), the auxiliary detection heads can better assist the model training. The reason may be that the coarse boxes contain a diversity of region samples, so that the auxiliary heads can facilitate the model training by taking advantage of the diversity.  


\noindent
{\bf Generalizability of DSFPN.}
To demonstrate the generalizability of the proposed DSFPN, we conduct extensive experiments on val2017 and test-dev datasets. The experimental results are listed in Table~\ref{table:generalizability}. By comparing the results in val2017 and test-dev, we note that in two-stage methods, the performance increment gained by adding DSFPN is present in all backbones; moreover, the improvement is consistent across the val2017 and test-dev. This demonstrates that the proposed DSFPN promotes detection performance not by over-fitting to a validation data, it can generalize to test dataset.
In multi-stage methods, the consistent improvement also exists, but with subtle less margin. Apart from detection task, in the instance segmentation, both two and multi-stage methods with DSFPN surpass that  without DSFPN.  The aforementioned observations illustrate that the proposed DSFPN is general and robust to datasets and tasks.

\noindent
{\bf Inference time.}
From Table~\ref{table:generalizability}, we see that in two-stage methods, with or without DSFPN, the inference speed is almost the same; in multi-stage methods, the speed has a trivial difference. The subtle change in speed demonstrates that our DSFPN strengthens detectors precision with little extra computational cost. 

\noindent
{\bf Visualization of feature maps.}
To verify that DSFPN does facilitate the feature learning in bottom-up pyramid and enhances the feature representation capability of top-down pyramid, we visualize the feature maps from each level of the pyramids in Fig.~\ref{fig:det_feat}. The displayed feature maps are the summation of the raw feature maps over channel at each pyramid level. 
We can note that the feature maps from each level of bottom-up pyramid of DSFPN (green boxes in right of Fig.~\ref{fig:det_feat}) are cleaner and with less noise than that from FPN~\cite{lin2017feature} (green boxes left of Fig.~\ref{fig:det_feat}). Accordingly, the feature maps from top-down pyramid (blue boxes in right of Fig.~\ref{fig:det_feat}) also have stronger activation, comparing with the counterpart (blue boxes in left of Fig.~\ref{fig:det_feat}). Especially, at lower levels, the feature maps from DSFPN are significantly superior to that from FPN~\cite{lin2017feature}.
These observations experimentally demonstrate that the proposed DSFPN guides feature learning more effectively than FPN~\cite{lin2017feature} does.

\vspace{-0.2cm}
\subsection{Comparison with state-of-the-arts}
To demonstrate the effectiveness of the  proposed DSFPN, it is compared with representative and recently proposed methods on the COCO test-dev dataset. The results are shown in Table~\ref{table:compare_sota}. In the ResNet101 setting, the proposed methods outperform all other one-, two-, or multi-stage methods. In multi-stage methods, the proposed method surpasses other methods by a large margin. These experimental results clearly show the advantage of the proposed method.
\subsection{Learning curves of FPN and DSFPN}
\begin{figure}[t]
	\begin{center}
	\includegraphics[width=.951\linewidth,height=.6\linewidth]{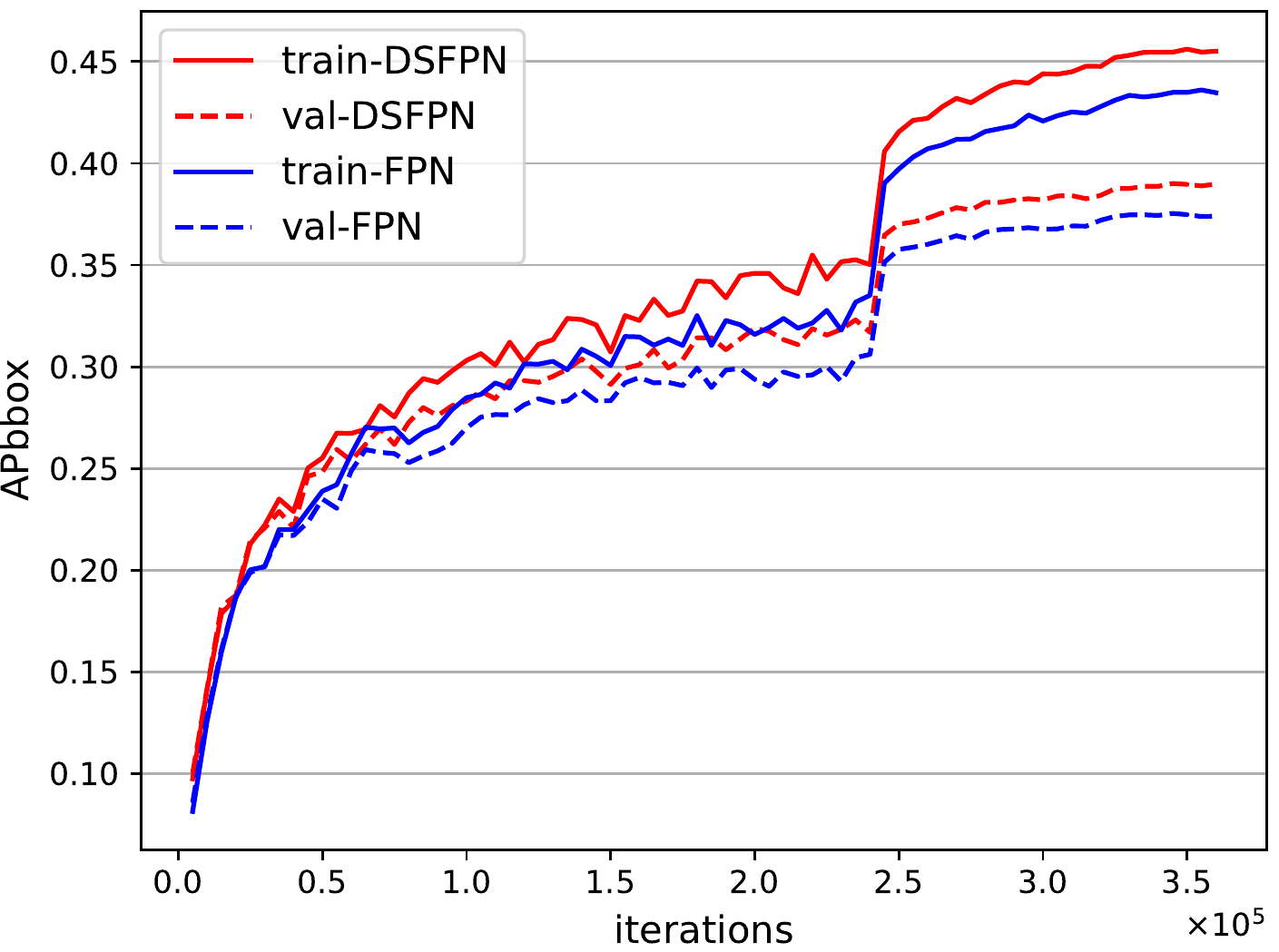}
	\end{center}
	\vspace{-5.5mm}
	\caption{Learning curves of detectors with FPN (blue) and DSFPN (red). The detector is Mask RCNN with ResNet50, trained in $1\times$ setting.}
	\label{fig:learning_curve}
\end{figure}

To study the effectiveness of the proposed dually supervised (DS) training strategy in facilitating model learning, we leverage Mask RCNN with ResNet50 as detector and compute the AP$_{bbox}$ over training data and validation data (val2017) with interval $5,000$ iterations in $1\times$ training setting. Due to the large volume of training data, we compute the metric over randomly sampled $10k$ images from entire training images.
From Fig.~\ref{fig:learning_curve}, we note that the detector with  DSFPN converges faster than that with FPN in training data (red solid line vs blue solid line). This can be attributed to DS training manner. Specifically, DS introduces auxiliary prediction heads to assist the final prediction heads via directly supervising bottom-up subnet. Therefore, the detector with DSFPN can converge more efficiently. Moreover, the detector with DSFPN can converge to a higher AP{$_{bbox}$} point. This demonstrates that DS not only promotes detector training efficiently but also effectively.
\section{Conclusion}
In this work, we propose a novel dually supervised training strategy, named DSFPN, to improve FPN~\cite{lin2017feature} by dealing with the gradient exploding or vanishing problem. DSFPN is generally effective for object detection and segmentation tasks, and it needs little extra computation during detection/segmentation inference. 
In addition, a simple but effective \textit{cls}-\textit{reg} decoupled head is proposed for object detection in DSFPN to alleviate task inconsistency. To demonstrate the superiority, generalizability, and efficiency of the proposed DSFPN, we integrate DSFPN into Faster/Mask/Cascade/CascadeMask RCNN, and promising results are observed in the extensive validation experiments on the COCO dataset.
\label{conclusion}

{\small
\bibliographystyle{ieee_fullname}
\bibliography{egbib}

\begin{thebibliography}{10}\itemsep=-1pt

\bibitem{cai2016unified}
Zhaowei Cai, Quanfu Fan, Rogerio~S Feris, and Nuno Vasconcelos.
\newblock A unified multi-scale deep convolutional neural network for fast
  object detection.
\newblock In {\em european conference on computer vision}, pages 354--370.
  Springer, 2016.

\bibitem{cai2018cascade}
Zhaowei Cai and Nuno Vasconcelos.
\newblock Cascade r-cnn: Delving into high quality object detection.
\newblock In {\em Proceedings of the IEEE conference on computer vision and
  pattern recognition}, pages 6154--6162, 2018.

\bibitem{cai2019cascade}
Zhaowei Cai and Nuno Vasconcelos.
\newblock Cascade r-cnn: High quality object detection and instance
  segmentation.
\newblock {\em arXiv preprint arXiv:1906.09756}, 2019.

\bibitem{cheng2018revisiting}
Bowen Cheng, Yunchao Wei, Honghui Shi, Rogerio Feris, Jinjun Xiong, and Thomas
  Huang.
\newblock Revisiting rcnn: On awakening the classification power of faster
  rcnn.
\newblock In {\em Proceedings of the European Conference on Computer Vision
  (ECCV)}, pages 453--468, 2018.

\bibitem{dai2016r}
Jifeng Dai, Yi Li, Kaiming He, and Jian Sun.
\newblock R-fcn: Object detection via region-based fully convolutional
  networks.
\newblock In {\em Advances in neural information processing systems}, pages
  379--387, 2016.

\bibitem{duan2019centernet}
Kaiwen Duan, Song Bai, Lingxi Xie, Honggang Qi, Qingming Huang, and Qi Tian.
\newblock Centernet: Object detection with keypoint triplets.
\newblock {\em arXiv preprint arXiv:1904.08189}, 2019.

\bibitem{fu2017dssd}
Cheng-Yang Fu, Wei Liu, Ananth Ranga, Ambrish Tyagi, and Alexander~C Berg.
\newblock Dssd: Deconvolutional single shot detector.
\newblock {\em arXiv preprint arXiv:1701.06659}, 2017.

\bibitem{girshick2015fast}
Ross Girshick.
\newblock Fast r-cnn.
\newblock In {\em Proceedings of the IEEE international conference on computer
  vision}, pages 1440--1448, 2015.

\bibitem{girshick2014rich}
Ross Girshick, Jeff Donahue, Trevor Darrell, and Jitendra Malik.
\newblock Rich feature hierarchies for accurate object detection and semantic
  segmentation.
\newblock In {\em Proceedings of the IEEE conference on computer vision and
  pattern recognition}, pages 580--587, 2014.

\bibitem{he2017mask}
Kaiming He, Georgia Gkioxari, Piotr Doll{\'a}r, and Ross Girshick.
\newblock Mask r-cnn.
\newblock In {\em Proceedings of the IEEE international conference on computer
  vision}, pages 2961--2969, 2017.

\bibitem{he2016deep}
Kaiming He, Xiangyu Zhang, Shaoqing Ren, and Jian Sun.
\newblock Deep residual learning for image recognition.
\newblock In {\em Proceedings of the IEEE conference on computer vision and
  pattern recognition}, pages 770--778, 2016.

\bibitem{law2018cornernet}
Hei Law and Jia Deng.
\newblock Cornernet: Detecting objects as paired keypoints.
\newblock In {\em Proceedings of the European Conference on Computer Vision
  (ECCV)}, pages 734--750, 2018.

\bibitem{lee2015deeply}
Chen-Yu Lee, Saining Xie, Patrick Gallagher, Zhengyou Zhang, and Zhuowen Tu.
\newblock Deeply-supervised nets.
\newblock In {\em Artificial intelligence and statistics}, pages 562--570,
  2015.

\bibitem{li2019scale}
Yanghao Li, Yuntao Chen, Naiyan Wang, and Zhaoxiang Zhang.
\newblock Scale-aware trident networks for object detection.
\newblock {\em arXiv preprint arXiv:1901.01892}, 2019.

\bibitem{lin2017feature}
Tsung-Yi Lin, Piotr Doll{\'a}r, Ross Girshick, Kaiming He, Bharath Hariharan,
  and Serge Belongie.
\newblock Feature pyramid networks for object detection.
\newblock In {\em Proceedings of the IEEE conference on computer vision and
  pattern recognition}, pages 2117--2125, 2017.

\bibitem{lin2017focal}
Tsung-Yi Lin, Priya Goyal, Ross Girshick, Kaiming He, and Piotr Doll{\'a}r.
\newblock Focal loss for dense object detection.
\newblock In {\em Proceedings of the IEEE international conference on computer
  vision}, pages 2980--2988, 2017.

\bibitem{lin2014microsoft}
Tsung-Yi Lin, Michael Maire, Serge Belongie, James Hays, Pietro Perona, Deva
  Ramanan, Piotr Doll{\'a}r, and C~Lawrence Zitnick.
\newblock Microsoft coco: Common objects in context.
\newblock In {\em European conference on computer vision}, pages 740--755.
  Springer, 2014.

\bibitem{liu2016ssd}
Wei Liu, Dragomir Anguelov, Dumitru Erhan, Christian Szegedy, Scott Reed,
  Cheng-Yang Fu, and Alexander~C Berg.
\newblock Ssd: Single shot multibox detector.
\newblock In {\em European conference on computer vision}, pages 21--37.
  Springer, 2016.

\bibitem{lu2019grid}
Xin Lu, Buyu Li, Yuxin Yue, Quanquan Li, and Junjie Yan.
\newblock Grid r-cnn.
\newblock In {\em Proceedings of the IEEE Conference on Computer Vision and
  Pattern Recognition}, pages 7363--7372, 2019.

\bibitem{pang2019libra}
Jiangmiao Pang, Kai Chen, Jianping Shi, Huajun Feng, Wanli Ouyang, and Dahua
  Lin.
\newblock Libra r-cnn: Towards balanced learning for object detection.
\newblock In {\em Proceedings of the IEEE Conference on Computer Vision and
  Pattern Recognition}, pages 821--830, 2019.

\bibitem{redmon2016you}
Joseph Redmon, Santosh Divvala, Ross Girshick, and Ali Farhadi.
\newblock You only look once: Unified, real-time object detection.
\newblock In {\em Proceedings of the IEEE conference on computer vision and
  pattern recognition}, pages 779--788, 2016.

\bibitem{redmon2017yolo9000}
Joseph Redmon and Ali Farhadi.
\newblock Yolo9000: better, faster, stronger.
\newblock In {\em Proceedings of the IEEE conference on computer vision and
  pattern recognition}, pages 7263--7271, 2017.

\bibitem{ren2015faster}
Shaoqing Ren, Kaiming He, Ross Girshick, and Jian Sun.
\newblock Faster r-cnn: Towards real-time object detection with region proposal
  networks.
\newblock In {\em Advances in neural information processing systems}, pages
  91--99, 2015.

\bibitem{ronneberger2015u}
Olaf Ronneberger, Philipp Fischer, and Thomas Brox.
\newblock U-net: Convolutional networks for biomedical image segmentation.
\newblock In {\em International Conference on Medical image computing and
  computer-assisted intervention}, pages 234--241. Springer, 2015.

\bibitem{simonyan2014very}
Karen Simonyan and Andrew Zisserman.
\newblock Very deep convolutional networks for large-scale image recognition.
\newblock {\em arXiv preprint arXiv:1409.1556}, 2014.

\bibitem{Tan_2019_ICCV}
Zhiyu Tan, Xuecheng Nie, Qi Qian, Nan Li, and Hao Li.
\newblock Learning to rank proposals for object detection.
\newblock In {\em The IEEE International Conference on Computer Vision (ICCV)},
  October 2019.

\bibitem{xie2017aggregated}
Saining Xie, Ross Girshick, Piotr Doll{\'a}r, Zhuowen Tu, and Kaiming He.
\newblock Aggregated residual transformations for deep neural networks.
\newblock In {\em Proceedings of the IEEE conference on computer vision and
  pattern recognition}, pages 1492--1500, 2017.

\bibitem{xie2015holistically}
Saining Xie and Zhuowen Tu.
\newblock Holistically-nested edge detection.
\newblock In {\em Proceedings of the IEEE international conference on computer
  vision}, pages 1395--1403, 2015.

\bibitem{yang2019reppoints}
Ze Yang, Shaohui Liu, Han Hu, Liwei Wang, and Stephen Lin.
\newblock Reppoints: Point set representation for object detection.
\newblock {\em arXiv preprint arXiv:1904.11490}, 2019.

\bibitem{zhang2018single}
Shifeng Zhang, Longyin Wen, Xiao Bian, Zhen Lei, and Stan~Z Li.
\newblock Single-shot refinement neural network for object detection.
\newblock In {\em Proceedings of the IEEE Conference on Computer Vision and
  Pattern Recognition}, pages 4203--4212, 2018.

\bibitem{zhao2017pyramid}
Hengshuang Zhao, Jianping Shi, Xiaojuan Qi, Xiaogang Wang, and Jiaya Jia.
\newblock Pyramid scene parsing network.
\newblock In {\em Proceedings of the IEEE conference on computer vision and
  pattern recognition}, pages 2881--2890, 2017.

\bibitem{zhu2019feature}
Chenchen Zhu, Yihui He, and Marios Savvides.
\newblock Feature selective anchor-free module for single-shot object
  detection.
\newblock {\em arXiv preprint arXiv:1903.00621}, 2019.

\end{thebibliography}
}
\end{document}